\def\year{2017}\relax
\documentclass[letterpaper]{article}
\usepackage{aaai17}
\usepackage{times}
\usepackage[table]{xcolor}
\usepackage{multirow,booktabs,array,xcolor,colortbl,multicol,epsfig,graphicx,subfigure,amsmath,amsthm,amsfonts,amssymb,bm,algorithm,algorithmic}
\usepackage{caption}
\usepackage{colortbl,accents}
\usepackage{subfigure}
\usepackage{caption}
\usepackage{bbding}
\renewcommand\footnotemark{}
 
\begin{document}
\title{Modeling The Intensity Function Of Point Process Via Recurrent Neural Networks}
\author{Shuai Xiao$^{1}$\thanks{Correspondence author is Junchi Yan. This research was partially supported by The National Key Research and Development Program of China (2016YFB1001003), NSFC (61602176, 61672231, 61527804, 61521062), STCSM (15JC1401700, 14XD1402100), China Postdoctoral Science Foundation Funded Project (2016M590337), the 111 Program (B07022) and NSF (IIS-1639792, DMS-1620345).}, Junchi Yan$^{23}$, \textbf{Stephen M. Chu}$^3$, \textbf{Xiaokang Yang}$^1$, \textbf{Hongyuan Zha}$^4$\\
$^1$ Shanghai Jiao Tong University \quad $^2$ East China Normal University \quad $^3$ IBM Research -- China \quad $^4$ Georgia Tech\\
 \{benjaminforever,xkyang\}@sjtu.edu.cn, jcyan@sei.ecnu.edu.cn, schu@us.ibm.com, zha@cc.gatech.edu
}
\maketitle
\begin{abstract}
Event sequence, asynchronously generated with random timestamp, is ubiquitous among applications. The precise and arbitrary timestamp can carry important clues about the underlying dynamics, and has lent the event data fundamentally different from the time-series whereby series is indexed with fixed and equal time interval. One expressive mathematical tool for modeling event is point process. The intensity functions of many point processes involve two components: the background and the effect by the history. Due to its inherent spontaneousness, the background can be treated as a time series while the other need to handle the history events. In this paper, we model the background by a Recurrent Neural Network (RNN) with its units aligned with time series indexes while the history effect is modeled by another RNN whose units are aligned with asynchronous events to capture the long-range dynamics. The whole model with event type and timestamp prediction output layers can be trained end-to-end. Our approach takes an RNN perspective to point process, and models its background and history effect. For utility, our method allows a black-box treatment for modeling the intensity which is often a pre-defined parametric form in point processes. Meanwhile end-to-end training opens the venue for reusing existing rich techniques in deep network for point process modeling. We apply our model to the predictive maintenance problem using a log dataset by more than 1000 ATMs from a global bank headquartered in North America.
\end{abstract}

\section{Introduction}%\footnote{One straightforward definition for time series can be found in http://www.itl.nist.gov/div898/handbook/pmc/section4/pmc41.htm (accessed on 8/8/2016): `\emph{An ordered sequence of values of a variable at equally spaced time intervals}'.}
Event sequence is becoming increasingly available in a variety of applications such as e-commerce transactions, social network activities, conflicts, and equipment failures etc. The event data can carry rich information not only about the event attribute (e.g. type, participator) but also the timestamp $\{z_i,t_i\}_{i=1}^N$ indicating when the event occurs. Being treated as a random variable when the event is stochastically generated in an asynchronous manner, the timestamp makes the event sequence fundamentally different from the time series \cite{MontgomeryTSBook15} with equal and fixed time interval, whereby the time point only serves as a role for index $\{y_t\}_{t=1}^T$. A major line of research \cite{AalenPP2008} has been devoted to study event sequence, especially exploring the timestamp information to model the underlying dynamics of the system, whereby point process \cite{SnyderPPBook12} has been a powerful and compact framework in this direction.

Recently there are many machine learning based models for scalable point process modeling. We attribute the progressions in this direction in part to the smart mathematical reformulations and optimization techniques e.g. \cite{LewisJNS2011,ZhouICML13,ZhouAISTATS13}, as well as novel parametric forms for the conditional intensity function \cite{shen2014modeling,ErtekinRPP2015,xu2016icu} as carefully designed by researchers' prior knowledge to capture the character of the dataset in study. However, one major limitation of the parametric forms of point process is due to their specialized and restricted expression capability for arbitrary distributed event data which trends to be oversimplified or even infeasible for capturing the problem complexity in real applications. Moreover, it runs at the risk of model underfitting due to the misjudgement on model choice. Recent works e.g. \cite{ZhouICML13} start to turn to non-parametric form to fit the structure of a point process, but their method is under the Hawkes process formulation,while this formulation runs at the risk of model mis-choice.

%While predictive models have been developed to anticipate needs, most existing work has focused on specialized predictive models with a narrow target.
In this paper, we view the conditional intensity of a point process as a nonlinear mapping between the predicted transient occurrence intensity of events with different types, and the model input information of event participators, event profile and the system history. Such a nonlinear mapping is expected to be complex and flexible enough to model various characters of real event data for its application utility.

In fact, deep learning models, such as Convolutional Neural Networks (CNNs) \cite{LeCunIEEE98}, Recurrent Neural Networks (RNNs) \cite{PascanuICML13} have attracted wide attention in recent vision, speech and language communities, and many of them has dominated the competing results on perceptual benchmark tasks e.g. \cite{ILSVRC15}. In particular, we turn to RNNs as a natural way to encode such nonlinear and dynamic mapping, in an effort for modeling an end-to-end nonlinear intensity mapping without any prior knowledge.
\begin{figure}[tb!]
	\centering
	\subfigure{\includegraphics[width=0.48\textwidth]{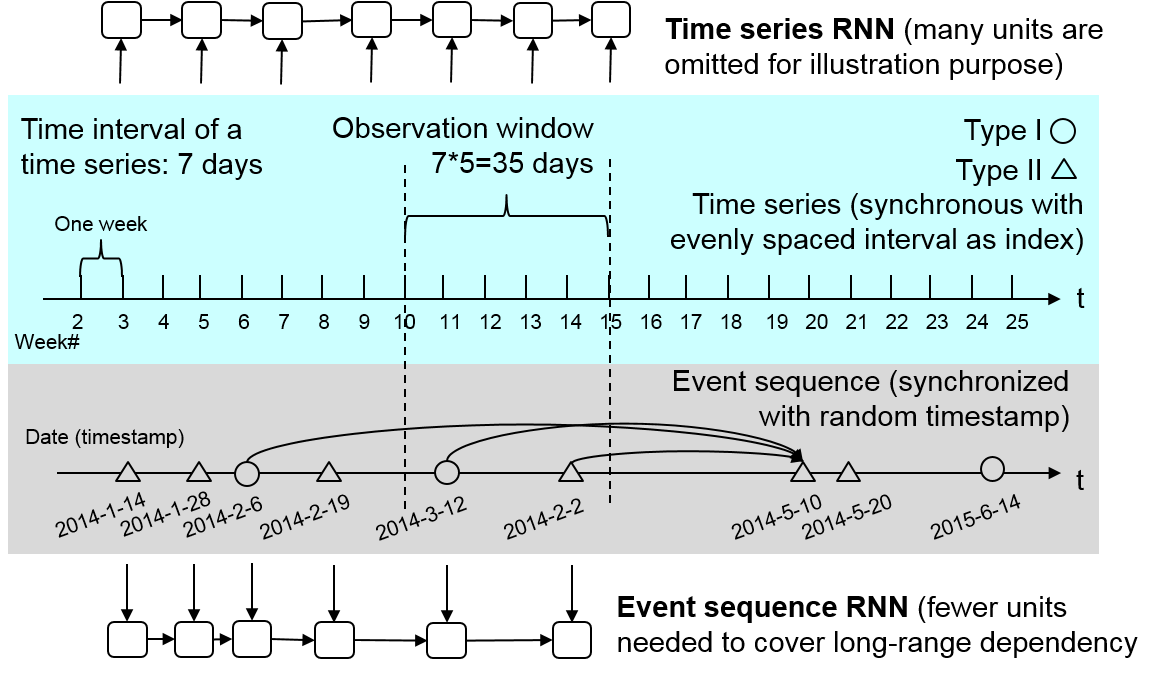}}%\vspace{-10pt}
	\caption{Time series and event sequence can be synergically modeled. The former can be used to timely capture the recent window for the time-varying features, while the latter can capture the long-range dependency over time. Note the dependency in an event sequence can be easily captured by an event sequence LSTM with less than 5 steps, while it takes too much more steps if using a time series with a fixed time interval e.g. 7 days in the figure (Note many unit steps in the top time series are omitted in the figure for clarity).}
	\label{fig:ppts}
\end{figure}

\textbf{Key idea and highlights} Our model interprets the conditional intensity function of a point process as a nonlinear mapping, which is synergetically established by a composite neural network with two RNNs as its building blocks. As illustrated in Fig.\ref{fig:ppts}, time series (top row) and event sequence (bottom row) are distinct to each other in that time series is more suitable to carry the synchronously (i.e. in a fixed pace) and regularly updated or constant profile features, while the event sequence can compactly catch event driven, more abrupt information, which can affect the condition intensity function over longer time period. More specifically, the highlights of this paper are:

1) We first make an observation that many conditional intensity functions can be viewed as an integration of two effects: i) spontaneous background component inherently affected by the internal (time-varying) attributes of the individual and the event type; ii) effects from history events. Meanwhile, most information in real world can also be covered by continuously updated features like age, temperature, and asynchronous event data such as clinical records, failures. This motivates us to devise a general approach.

2) Then we use an RNN whose units are aligned with the time points of a time series as its units, and an RNN whose units are aligned with events. The time series RNN can timely track the spontaneous background while the event sequence RNN is used to efficiently capture the long-range dependency over history with arbitrary time intervals. This allows to fit arbitrary dynamics of point process which otherwise will be very difficult or often impossible to be specified by a parameterized model under certain assumptions.

3) To our best knowledge, this is the first work to \emph{fully} interpret and instantiate the conditional intensity function with fused time series and event sequence RNNs. This opens up the room for connecting the neural network techniques to traditional point process that emphasizes more on specific model driven by domain knowledge. More importantly the introduction of a full RNN treatment lessen the efforts for the design of (semi-)parametric point process model and its complex learning algorithms which often call for special tricks that prohibiting the wide use for practitioners. In contrast, neural networks and specifically RNN, is becoming off-the-shelf tools and getting widely used recently.

4) Our model is simple, general and can be end-to-end trained. We target to a predictive maintenance problem. The data is from a global bank headquartered in North America, consisting decades of thousands of event logs for a large number of Automated Teller Machines (ATMs). The state-of-the-art performance on failure type and timestamp prediction corroborates its suitability to real-world applications.

%between the input attributes and timestamp associated with individuals and their events
\section{Related Work and Motivation}
We view the related concepts and work in this section, which is mainly focused on Recurrent Neural Networks (RNNs) and their applications in time series and sequences data, respectively. Then we give our point of view on existing point process methods and their connection to RNNs. All these observations indeed motivate the work of this paper.

\textbf{Recurrent neural network} The building blocks of our model are the Recurrent Neural Networks (RNNs) \cite{ElmanCS90,PascanuICML13} and its modern variant Long Short-Term Memory (LSTM) units \cite{HochreiterNC97,GravesArxiv13}. RNNs are dynamical systems whose next state and output depend on the present network state and input, which are more general models than the feed-forward networks. RNNs have long been explored in perceptual applications for many decades, however it can be very difficult to train RNNs to learn long-range dynamics perhaps in part due to the vanishing and exploding gradients problem. LSTMs provide a solution by incorporating memory units that allow the network to learn when to forget previous hidden states and when to update hidden states given new information. Recently, RNNs and LSTMs have been successfully applied in large-scale vision \cite{GregorDraw15}, speech \cite{GravesICML14} and language \cite{SutskeverNIPS14} problems.

\textbf{RNNs for series data} From application perspective, we view RNNs works by two scenarios as particularly considered in this paper: i) RNNs for synchronized series with evenly spaced interval e.g. time series or indexed sequence with pure order information e.g. language; ii) asynchronous sequence with random timestamp e.g. event data.

\emph{i) Synchronized series}: RNNs have been a long time a natural tool for standard time series modeling and prediction \cite{ConnorTNN94,HanTSP04,ChandraNC12,ChenJH13}, whereby the indexed series data point is fed as input to an (unfold) RNN. In a broader sense, video frames can also be treated as time series and RNN are widely used in recent visual analytics works \cite{JainICRA16,TripathiBMVC16} and so for speech \cite{GravesICML14}. RNNs are also intensively adopted for sequence modeling tasks \cite{ChungArxiv14,BengioNIPS15} when only order information is considered.

\emph{ii) Asynchronous event}: In contrast, event sequence with timestamp about their occurrence, which are asynchronously and randomly distributed over the continuous time space, is another typical input type for RNNs \cite{DuKDD16,ChoiArxiv16,EstebanArxiv16} and \cite{CheArxiv16} (despite its title for 'time series'). One key differentiation against the first scenario is that the timestamp or time duration between events (together with other features) is taken as input to the RNNs. By doing so, (long-range) event dependency can be effectively encoded.

\textbf{Point process} Point process has been a principled framework for modeling event data \cite{AalenPP2008}. The dynamics of the point process can be well captured by its conditional intensity function whose definition is briefly reviewed here: for a short time window $[t,t+dt)$, $\lambda(t)$ represents the rate for the occurrence of a new event conditioned on the history $\mathcal{H}_t = \{z_i,t_i|t_i < t\}$:
\begin{equation}\notag
\lambda(t)=\lim_{\Delta t\rightarrow 0}\frac{\mathbb{E}(N(t+\Delta t)-N(t)|\mathcal{H}_t)}{\Delta t}=\frac{\mathbb{E}(dN(t)|\mathcal{H}_t)}{dt}
\end{equation}
where $\mathbb{E}(dN(t)|\mathcal{H}_t)$ is the expectation of the number of events happened in the interval $(t, t + dt]$ given the historical observations $\mathcal{H}_t$. The conditional intensity function has played a central role in point processes and many popular processes vary on how it is parameterized.

1) \emph{Poisson process} \cite{KingmanPP92}: the homogeneous Poisson process has a very simple form for its intensity function: $\lambda(t)=\lambda_{0}$. Poisson process and its time-varying generalization are both assumed to be independent of the history.

2) \emph{Reinforced poisson processes} \cite{PemantlePS07,shen2014modeling}: the model captures the `rich-get-richer' mechanism characterized by a compact intensity function, which is recently used for popularity prediction \cite{shen2014modeling}.

3) \emph{Hawkes process} \cite{HawkesBiometrika71}: Hawkes process has received wide attention recently in social network analysis \cite{ZhouAISTATS13}, viral diffusion\cite{yang2013mixture} and criminology \cite{lewis2010self} etc. It explicitly uses a triggering term to model the excitation effect from history events and is originally motivated to analyze the earthquake and its aftershocks\cite{OgataJASA88}.

4) \emph{Reactive point process} \cite{ErtekinRPP2015}: it can be regarded as a generalization for the Hawkes process by adding a self-inhibiting term to account for the inhibiting effects from history events.

5) \emph{Self-correcting process} \cite{IshamSPP79}: its background part increases steadily, while it is decreased by a constant $e^{-\alpha}<1$ every time a new event appears.

We reformulate these intensity functions in their general form in Table \ref{tab:intensity}. It tries to separate the spontaneous background component and history event effect explicitly.

\textbf{Predictive maintenance} Predictive maintenance \cite{MobleyBH02} is a sound testbed for our model which refers to a practice that involves equipment risk prediction to allow for proactive scheduling of corrective maintenance. Such an early identification of potential concerns helps deploy limited resources more cost effectively, reduce operations costs and maximize equipment uptime \cite{GrallTR02}. Predictive maintenance is adopted in a wide variety of applications such as fire inspection \cite{MadaioKDD16}, data center \cite{SirbuPPW15} and electrical grid \cite{ErtekinRPP2015} management. For its practical importance in different scenarios and relative rich event data for modeling, we target our model to a real-world dataset of more than 1,000 automated teller machines (ATMs) from a global bank headquartered in North America.

\begin{table}[tb!]
\centering
\caption{Conditional intensity functions of point processes.}
\resizebox{0.48\textwidth}{!}{
\begin{tabular}{lrr}
  \addlinespace
  \toprule
  Model&Background&History event effect\\
  \midrule
  Poisson process&$\mu(t)$&$0$\\
  Reinforced poisson process&$0$&$\gamma(t)\sum_{t_i<t}\delta(t_i<t)$\\
  %Hawkes process&$\lambda_0$&$\sum_{t_i<t}\alpha \omega\exp(-\omega(t-t_i)),\omega>0$\\
  Hawkes process&$\mu(t)$&$\sum_{t_i<t}\gamma(t,t_i)$\\
  Reactive point process&$\mu(t)$&$\sum_{t_i<t}\gamma_1(t,t_i)-\sum_{t_i<t}\gamma_2(t,t_i)$\\
  Self-correcting process&$0$&$\exp(\mu t-\sum_{t_i<t}\gamma(t,t_i)$\\
  \bottomrule
\end{tabular}}
\small{Note:$\delta(t)$ is Dirac function, $\gamma(t,t_i)$ is time-decaying kernel and $\mu(t)$ can be constant or time-varying function.}
\label{tab:intensity}
\end{table}
\section{Network Structure and End-to-End Learning}
%\subsection{Brief on RNN as building block}
 Taking a sequence $\{\textbf{x}\}_{t=1}^T$ as input, the RNN generates the hidden states $\{\textbf{h}\}_{t=1}^T$ and outputs a sequence \cite{ElmanCS90,PascanuICML13}.
%\begin{align}\notag
%\textbf{h}_t&= f(\textbf{W}\textbf{x}_t+\textbf{H}\textbf{h}_{t-1}+\textbf{b}),\\\notag
%\textbf{y}_t&= \text{softmax}(\textbf{W}\textbf{x}_t+\textbf{b}_y)
%\end{align}%\label{eq-rnn}
Specifically, we implement our RNN with Long Short Term Memory (LSTM) \cite{HochreiterNC97,GravesArxiv13} for its popularity and well-known capability for efficient long-range dependency learning. In fact other RNN variant e.g. Gated Recurrent Units (GRU) \cite{ChungArxiv14} can also be alternative choice. We reiterate the formulation of LSTM:
\begin{align}\notag
\textbf{i}_t&= \sigma(\textbf{W}_i\textbf{x}_t+\textbf{U}_i\textbf{h}_{t-1}+\textbf{V}_i\textbf{c}_{t-1}+\textbf{b}_{i}),\\\notag
\textbf{f}_t&= \sigma(\textbf{W}_f\textbf{x}_t+\textbf{U}_f\textbf{h}_{t-1}+\textbf{V}_f\textbf{c}_{t-1}+\textbf{b}_{f}),\\\notag
\textbf{c}_t&= \textbf{f}_t\textbf{c}_{t-1}+\textbf{i}_t\odot\text{tanh}(\textbf{W}_c\textbf{x}_t+\textbf{U}_c\textbf{h}_{t-1}+\textbf{b}_c),\\\notag
\textbf{o}_t&= \sigma(\textbf{W}_o\textbf{x}_t+\textbf{U}_o\textbf{h}_{t-1}+\textbf{V}_o\textbf{c}_{t}+\textbf{b}_{o}),\\\notag
\textbf{h}_t&=\textbf{o}_t\odot\text{tanh}(\textbf{c}_t)
\end{align}%\label{eq-lstm}
where $\odot$ denotes element-wise multiplication and the recurrent activation $\sigma$ is the Logistic Sigmod function.
The above system can be reduced into an LSTM equation:
%One can see clearly that the nonlinear function $f$ in the standard RNN can be replaced by the LSTM equations:
\begin{equation}\notag
(\textbf{h}_t,\textbf{c}_t)= \text{LSTM}(\textbf{x}_t,\textbf{h}_{t-1}+\textbf{c}_{t-1})
\end{equation}%\label{eq-rnn}
%\subsection{Network structure and end-to-end learning}
We consider two types of input: i) continuously and evenly distributed time-series data e.g. temperature; ii) event data whose occurrence time interval is random. The network is comprised by two RNNs using evenly spaced time series $\{y_t\}_{t=1}^T$ to model the background intensity of events occurrence and event sequence $\{z_i,t_i\}_{i=1}^N$ to capture long-range event dependency. As a result, we have:
\begin{align}
(\textbf{h}^y_t,\textbf{c}^y_t)&= \text{LSTM}_{y}(\textbf{y}_t,\textbf{h}^y_{t-1}+\textbf{c}^y_{t-1}),\\
(\textbf{h}^z_t,\textbf{c}^z_t)&= \text{LSTM}_{z}(\textbf{z}_t,\textbf{h}^z_{t-1}+\textbf{c}^z_{t-1}),\\
\textbf{e}_t &= \text{tanh}(\textbf{W}_f[\textbf{h}^y_t,\textbf{h}^z_t]+\textbf{b}_f),\\
\textbf{U}_t &=\text{softMax}(\textbf{W}_U\textbf{e}_t+\textbf{b}_U),\\
\textbf{u}_t &=\text{softMax}(\textbf{W}_u[\textbf{e}_t,\textbf{U}_t]+\textbf{b}_u)\\
{s}_t &= \textbf{W}_s\textbf{e}_t+{b}_s,
\end{align}%\label{eq-lstm}
where $U$ and $u$ denotes the main type and subtype of events respectively. $s$ is the timestamp associated with each event. The total loss is the sum of the time prediction loss and the cross-entropy loss for event type:
\begin{footnotesize}
\begin{equation}
\sum_{j=1}^N\left(-W_U^j\log(U^j_{t})-w_u^j\log(u^j_{t})-\log\left(f(s^j_t|h^j_{t-1})\right)\right)
\label{eq:loss}
\end{equation}
\end{footnotesize}
where $N$ is the number of training samples indexed by $j$, and $s^j_t$ is the timestamp for the coming event, while $h^j_{t-1}$ is the history information. The underlying rationale for the third term is that we not only encourage correct classification of the coming event type, but also reinforce the corresponding timestamp of the event shall be close to the ground truth. We adopt a Gaussian penalty function with a fixed $\sigma^2=10$:
$$
f(s^j_t|h^j_{t-1})=\frac{1}{\sqrt{2\pi\sigma}}\exp\left(\frac{-(s^j_{t}-\tilde{s}^j_{t})^2}{2\sigma^2}\right)
$$
The output $\tilde{s}^j_{t}$ from the timestamp prediction layer is fed to the classification loss layer to compute the above penalty given the actual timestamp $s^j_{t}$ for sample $i$.

Following the importance weighting methodology for skewed data of model training \cite{rosenberg2012classifying}, the weight parameters $W, w$ for both main-type and subtype are set as the inverse of the sample number ratio in that type against the total size of samples, in order to weight more on those classes with fewer training samples. For the loss of independent main-type or subtype prediction as shown in Fig.\ref{fig:flat_pred}, we set the weight parameter $w$ and $W$ to zero respectively.
\begin{figure}[tb!]
	\centering
	\subfigure{\includegraphics[width=0.48\textwidth]{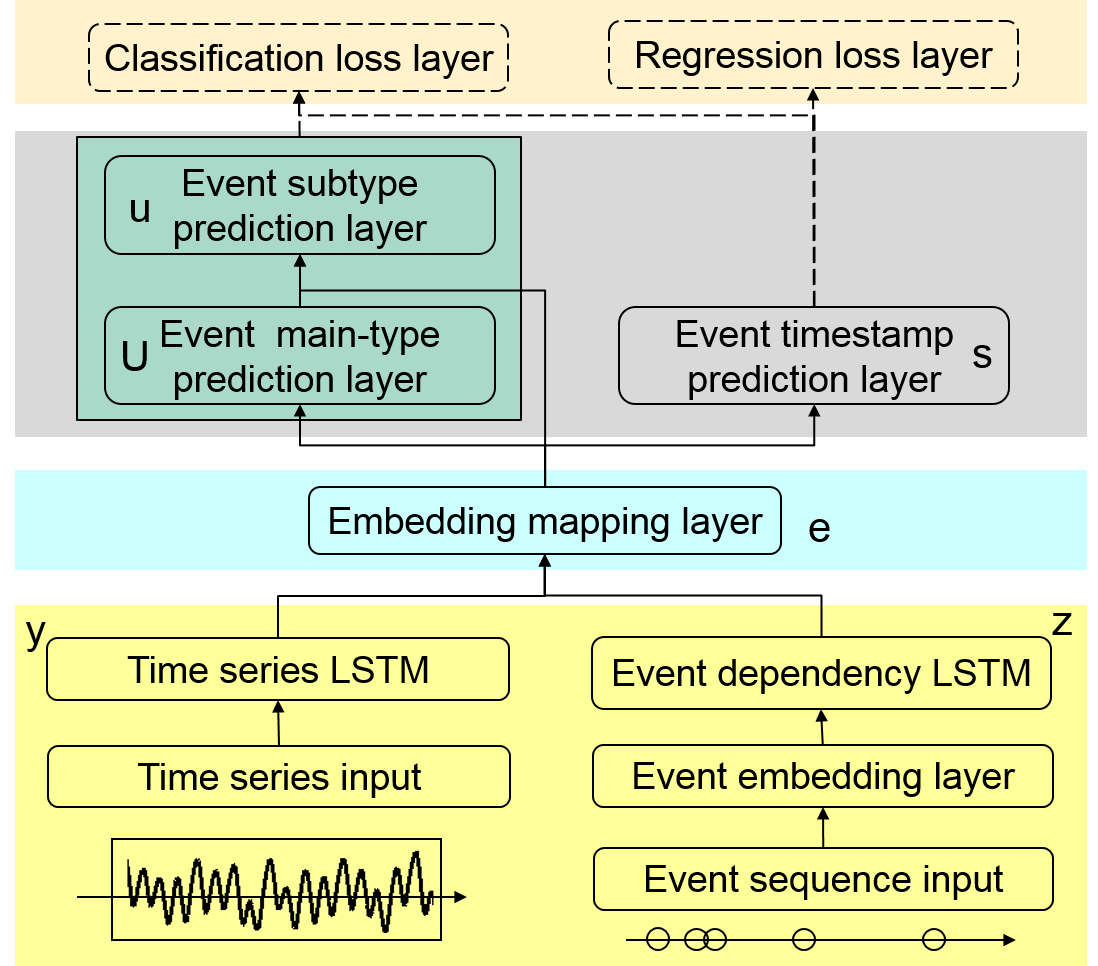}}%\vspace{-10pt}
	\caption{Our network can be trained end-to-end. Time series and event sequence are fed into two RNNs (LSTM) which are connected to an embedding mapping layer that fuses the information from two LSTMs. Then three prediction layers are used to output the predicted main type, subtype of events, and the associated timestamp. Cross-entropy with the time penalty loss by Eq.\ref{eq:loss} and square loss are respectively used for event type and timestamp prediction.}
	\label{fig:overview}
\end{figure}

We adopt RMSprop gradients \cite{RmspropArxiv15} which have been shown to work well on training deep networks to learn these parameters. %The step size is set to $0.001$.
\section{Experiments on Real-world Data}
We use failure prediction for predictive ATMs maintenance as a typical example of event based point process modeling. We have no prior knowledge on the dynamics of the complex system and the task can involve arbitrarily working schedules and heterogeneous mix of conditions. It takes much cost or even impractical to devise specialized models.
\begin{table}[tb!]
\centering
\caption{Statistics of main/sub-type of event count per ATM, and timestamp interval in days for all ATMs (in brackets).}%
\resizebox{0.48\textwidth}{!}{
\begin{tabular}{clrrrrr}
  %\addlinespace
  \toprule
data&type&total&max&min&mean&std\\
  \midrule
  &Ticket&2226(--)&10(137.04)&0(1.21)&2.09(31.70)&1.85(25.14)\\
  &Error&28434(--)&168(153.90)&0(0.10)&26.70(6.31)&18.38(9.74)\\
  \multirow{2}{*}{\rotatebox{90}{Training set}}&PRT&9204(--)&88(210.13)&0(0.10)&8.64(12.12)&11.37(21.41)\\
  &CNG&7767(--)&50(200.07)&0(0.10)&7.29(15.49)&6.59(23.87)\\
  &IDC&4082(--)&116(206.61)&0(0.10)&3.83(23.85)&5.84(30.71)\\
  &COMM&3371(--)&47(202.79)&0(0.10)&3.16(22.35)&3.90(29.36)\\
  &LMTP&2525(--)&81(207.93)&0(0.10)&2.37(22.86)&4.41(34.56)\\
  &MISC&1485(--)&32(204.41)&0(0.10)&1.39(24.27)&2.54(34.38)\\
  \midrule
  &Ticket&1164(--)&15(148.00)&0(0.13)&2.52(26.30)&2.41(25.22)\\
  &Error&11799(--)&104(193.75)&1(0.10)&25.59(6.47)&17.71(10.09)\\
  \multirow{2}{*}{\rotatebox{90}{Testing set}}&PRT&4089(--)&60(205.48)&0(0.10)&8.86(11.45)&10.77(20.44)\\
  &CNG&3134(--)&66(196.93)&0(0.10)&6.79(15.90)&6.91(25.22)\\
  &IDC&1645(--)&35(205.75)&0(0.10)&3.57(25.15)&4.29(31.62)\\
  &COMM&1366(--)&53(205.17)&0(0.10)&2.96(22.88)&4.08(30.09)\\
  &LMTP&939(--)&20(186.87)&0(0.10)&2.04(26.63)&2.76(36.96)\\
  &MISC&626(--)&21(190.75)&0(0.10)&1.36(25.14)&2.59(34.91)\\
  \bottomrule
\end{tabular}}
\label{tab:type_statistics}
\end{table}
\subsection{Problem and real data description}
In maintenance support services, when a device fails, the equipment owner raises a maintenance service ticket and technician will be assigned to repair the failure. In fact, the history log and relevant profile information about the equipment can be indicative signals for the coming failures.

The studied dataset is comprised of the event logs involving error reporting and failure tickets, which is originally collected from a large number of ATMs owned by an anonymous global bank headquartered in North America. The bank is also a customer of the technical support service department of a Fortune 500 IT company.

\textbf{ATM models} The training data consists of 1085 ATMs and testing data has 469 ATMs, in total 1557 Wincor ATMs that cover 5 ATM machine models: ProCash 2100 RL (980, 430), 1500 RL (19, 5), 2100 FL (53, 21), 1500 FL (26, 10), and 2250XE RL (7, 3). The numbers in the bracket indicate the number of machines for training and testing.

%The event log i.e. error records include device identity, timestamp, message content, priority, code, and action.
\textbf{Event type} There are two main types `ticket' and `error' from Sep. 2014 to Mar. 2015. Statistics is presented in Table \ref{tab:type_statistics}. Moreover `error' is divided into 6 subtypes regarding in which component the error occurs:
1) printer (PRT), 2) cash dispenser module (CNG), 3) internet data center (IDC), 4) communication part (COMM), 5) printer monitor (LMTP), 6) miscellaneous e.g. hip card module, usb (MISC).

\textbf{Features} The input features for the two RNNs are: \textbf{1)} \emph{Time series RNN}: For each sub-window of length 7 days, for the time series RNN, we extract features including: i) the inventory information: ATM models, age, location, etc; ii) Event statistics, including tickets events from maintenance records, and errors from system log. Their occurrence frequencies are used as features. The concatenation of the above two categories of features serves as the features for each sub-window i.e. time series point. \textbf{2)} \emph{Event sequence RNN}: event type and the time interval between two events.

\textbf{Model setting} We use a single layer LSTM of size 32 with Sigmoid gate activations and tanh activation for hidden representation. The embedding layer is fully connected and it uses tanh activation and outputs a 16 dimensional vector. One-hot or embedding can be used for event type representation. For a large number of types, embedding representation is compact and efficient. For time series RNN, we set the length of each sub-window (i.e. the evenly spaced time interval) to be 7 days and the number of sub-window to be 5. In this way, our observation length is 35 days for time series. For event-dependency, the length of event sequence can be arbitrarily long. Here we take it by 7.

We also test degraded versions of our model as follows:

\textbf{1)} \textbf{Time series RNN}: the input is event sequence (the right half in the yellow part of Fig.\ref{fig:overview}) is removed. Note this design is in spirit similar to many LSTM models \cite{JainICRA16,TripathiBMVC16} used for video analytics, whereby the frame sequence can be treated as time series as the input to LSTMs. \textbf{2)} \textbf{Event (sequence) RNN}: the RNN whose input is time series (the left half in the yellow part of Fig.\ref{fig:overview}) is removed; \textbf{3)} \textbf{Intensity RNN}: two RNN are fused as shown in Fig.\ref{fig:overview}. For the above three methods, the output layer is directly the fine-grained subtype of events with no hierarchical structure as shown in the top left part of Fig.\ref{fig:overview}) in dark green. We also term three `hierarchical' versions whose two hierarchical prediction layers in Fig.\ref{fig:overview} are used: \textbf{4)} \textbf{Time series hRNN}, \textbf{5)} \textbf{Event (sequence) hRNN}, \textbf{6)} \textbf{Intensity hRNN}.

In addition, we compare three major peer methods. For Logistic model, the input are the concatenation of feature vectors for all active time series RNN sub-windows (set to 5 in this paper). For RMTPP and Hawkes process, we train the model on the event sequences with associated information. In fact, RMTPP will further process the event data into the similar input information to our event RNN.

1) \textbf{Logistic model}: we use Logistic regression for event timestamp prediction and use another independent Logistic classification model for event type prediction.

2) \textbf{Recurrent Marked Temporal Point Processes (RMTPP)}: \cite{DuKDD16} uses neural network to model the event dependency flexibly. The method can only sample transient time series features when an event happens and use partially parametric form for the base intensity.

3) \textbf{Hawkes Process}: To enable multi-type event prediction, we use a Multi-dimensional Hawkes process. Similar to \cite{ZhouAISTATS13}, we also add a sparsity regularization term on the mutual infection matrix but the low-rank assumption is removed as we only have 6 subtypes.

\textbf{Evaluation metrics} We use several popular prediction metrics for performance evaluation. For the coming event type prediction, we adopt \emph{Precision}, \emph{Recall}, \emph{F1 Score} and \emph{Confusion matrix} over 2 main types (`error', `ticket') as well as Confusion matrix over 6 subtypes under `error'. Note all these metrics are computed for each type, and then are averaged over all types. For event time prediction, we use the \emph{Mean Absolute Error (MAE)} which measures the absolute difference between the predicted time point and the actual one. These settings are similar to \cite{DuKDD16}.

To evaluate the type and timestamp prediction jointly, we devise two more strict metrics. For type prediction, we narrow down to test the samples whose timestamp prediction error \emph{MAE}$<3$ days and we compute the new \emph{F1 score+}. For timestamp, we recompute the new \emph{MAE+} only for the samples whose coming event is correctly predicted.

\textbf{Platform} The code is based on Theano running on a Linux server with 32G memory, 2 CPUs with 6 cores for each: Intel(R) Xeon(R) CPU E5-2603 v3 @ 1.60GHz. We also use 4 GPU:GeForce GTX TITAN X for acculturation.

\begin{figure}[tb!]
\centering
  \subfigure{\label{fig:hier_pred}
  \includegraphics[width=0.16\textwidth]{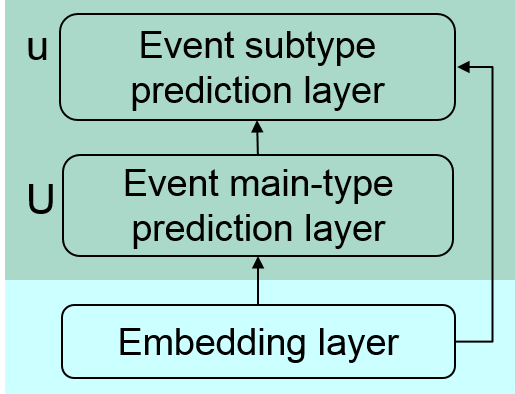}}%\hspace{-2pt}
  \subfigure{\label{fig:flat_pred}
  \includegraphics[width=0.3045\textwidth]{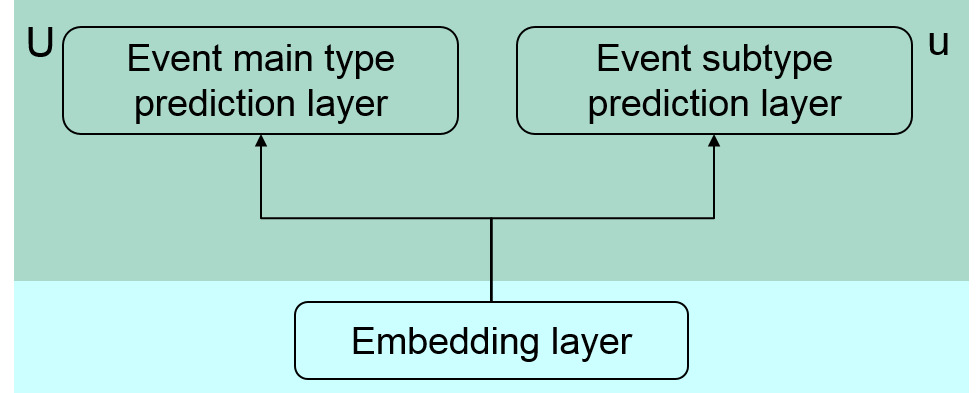}}%\vspace{-10pt}
	\caption{Hierarchical layer and Flat independent layer.}
	\label{fig:hierarchical}
\end{figure}
\begin{table}[tb!]
\centering
\caption{Ablation test of our method and peer methods i.e. multi-dimensional Hawkes process, recurrent Hawkes process and Logistic classification (for type) and regression (for event timestamp). Numbers are averaged over types.}
\resizebox{0.48\textwidth}{!}{
\begin{tabular}{clrrrr}
 \toprule
 &\multirow{2}{*}{model} &\multirow{2}{*}{main-type}&\multirow{2}{*}{subtype}&
 \multicolumn{2}{c}{hierarchical output} \\
  &&&&main-type & subtype \\
 \midrule
  &Time series RNN & 0.673 & 0.554 & 0.582 & 0.590\\
  \multirow{3}{*}{\rotatebox{90}{precision}}&Event sequence RNN &0.671 &0.570 &0.623 &0.614\\
  &\textbf{Intensity RNN} & 0.714&\textbf{0.620}&\textbf{0.642}&\textbf{0.664}\\
  &Hawkes process & 0.457&0.387& --- & ---- \\
  &Logistic prediction &\textbf{0.883}&0.385 & --- & --- \\
  &RMTPP& 0.581&0.574 & ---&---\\
 \midrule
 &Time series RNN &0.853&0.522&0.738&0.608 \\
  &Event sequence RNN & 0.821&0.543&0.770&0.621 \\
  \multirow{3}{*}{\rotatebox{90}{recall}}&Intensity RNN &\textbf{0.905}&\textbf{0.614}&\textbf{0.805}&\textbf{0.661}\\
  &Hawkes process & 0.493&0.394 &--- & --- \\
  &Logistic prediction & 0.795&0.273 &--- &--- \\
  &RMTPP& 0.691&0.583 &--- & --- \\
  \midrule
 &Time series RNN &0.707&0.533&0.571&0.605 \\
   \multirow{3}{*}{\rotatebox{90}{F1 score}}&Event sequence RNN & 0.703&0.555&0.651&0.610 \\
 &\textbf{Intensity RNN} & 0.765&\textbf{0.616}&\textbf{0.662}&\textbf{0.663}\\
  &Hawkes process &0.473&0.386& --- & ---\\
  &Logistic prediction &\textbf{0.832}&0.269 & --- & --- \\
  &RMTPP& 0.584&0.572 & --- & ---\\
  \midrule
  \multirow{3}{*}{\rotatebox{90}{MAE (in days)}}&Time series RNN & 4.37&4.48&4.26&4.41 \\
  &Event sequence RNN & 4.24&4.42&4.21&4.37 \\
  &\textbf{Intensity RNN} & \textbf{4.13}&\textbf{4.20}&\textbf{4.02}&\textbf{4.13} \\
  &Hawkes process & 5.26&5.46 & --- & --- \\
  &Logistic prediction & 4.52&4.61 & --- & --- \\
  &RMTPP& 4.28&4.32 & --- &--- \\
  \midrule
  &Time series RNN & 0.768&0.547&0.572&0.603 \\
  \multirow{3}{*}{\rotatebox{90}{{F1 Score+}}}&Event sequence RNN & 0.705&0.597&0.639&0.646\\
  &\textbf{Intensity RNN} & 0.825&\textbf{0.661}&\textbf{0.684}&\textbf{0.708} \\
  &Hawkes process & 0.467&0.451 & --- & --- \\
  &Logistic prediction & \textbf{0.846}&0.286 & --- & --- \\
  &RMTPP& 0.584&0.619 & --- &--- \\
  \midrule
  &Time series RNN & 4.21&3.78&\textbf{4.05}&\textbf{3.97}\\
  &Event sequence RNN & 4.16&3.84&4.12&4.01 \\
  \multirow{3}{*}{\rotatebox{90}{{MAE+}}}&\textbf{Intensity RNN} & \textbf{4.12}&\textbf{3.57}&4.21&4.11\\
  &Hawkes process & 5.42&3.93 & --- & --- \\
  &Logistic prediction & 4.5&4.24 & --- & --- \\
  &RMTPP& 4.26&3.99 & --- &--- \\
 \bottomrule
 \end{tabular}}
\label{tab:performance}
\end{table}

\begin{figure}[htb!]
\centering
  \subfigure[\scriptsize Intensity hRNN]{\label{fig:hiRNN_confusion}
  \includegraphics[width=0.145\textwidth]{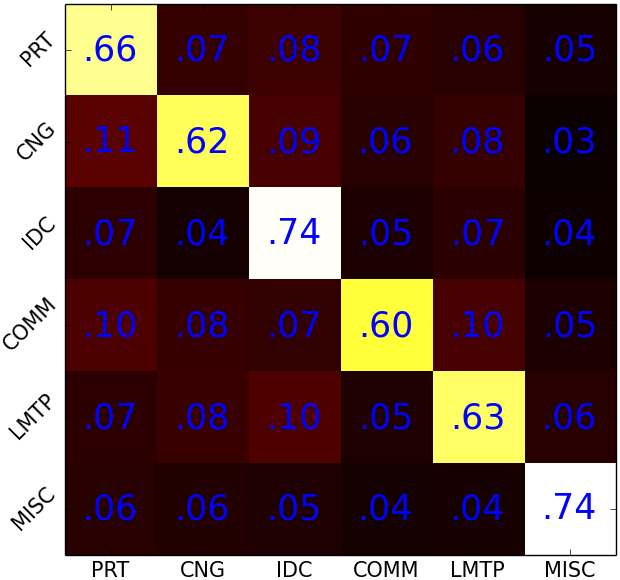}}%\hspace{-2pt}
  \subfigure[\scriptsize Time series hRNN]{\label{fig:htRNN_confusion}
  \includegraphics[width=0.145\textwidth]{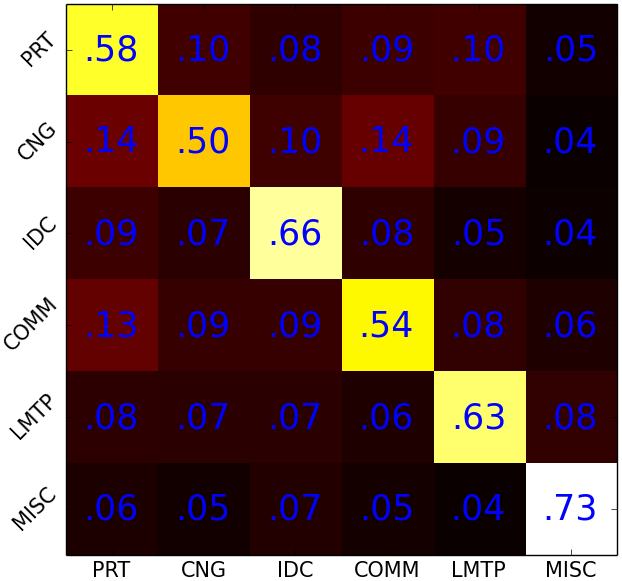}}%\hspace{-2pt}
  \subfigure[\scriptsize Event hRNN]{\label{fig:heRNN_confusion}
  \includegraphics[width=0.145\textwidth]{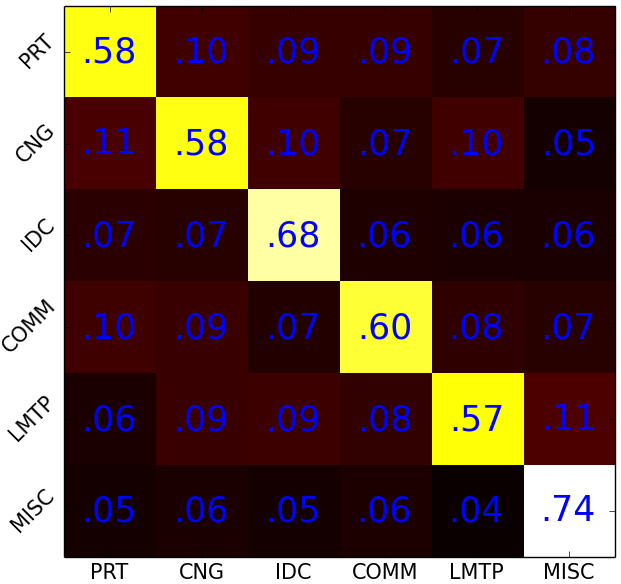}}\\%\vspace{-8pt}
  \subfigure[\scriptsize Intensity RNN]{\label{fig:iRNN_confusion}
  \includegraphics[width=0.145\textwidth]{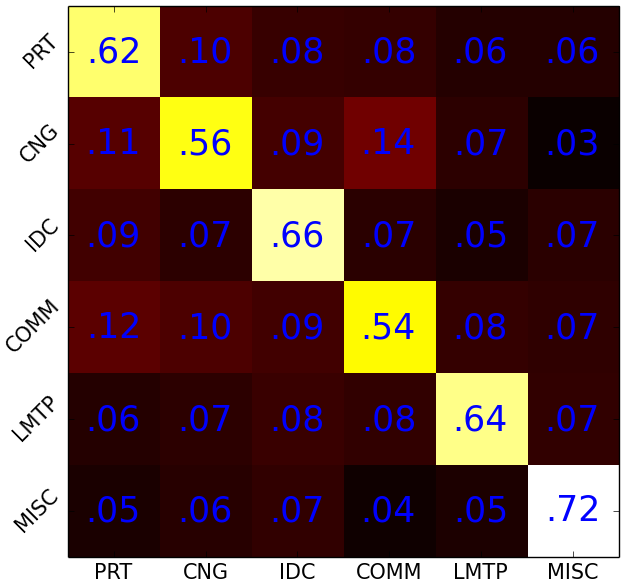}}%\hspace{-2pt}
  \subfigure[\scriptsize Time series RNN]{\label{fig:tRNN_confusion}
  \includegraphics[width=0.145\textwidth]{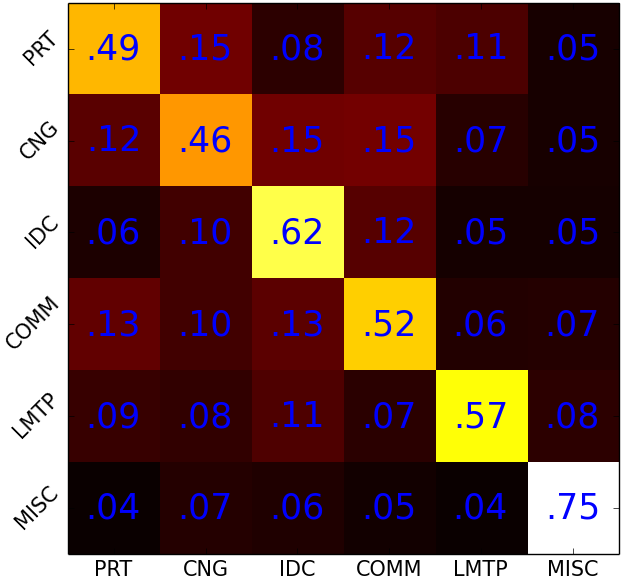}}%\hspace{-2pt}
  \subfigure[\scriptsize Event RNN]{\label{fig:eRNN_confusion}
  \includegraphics[width=0.145\textwidth]{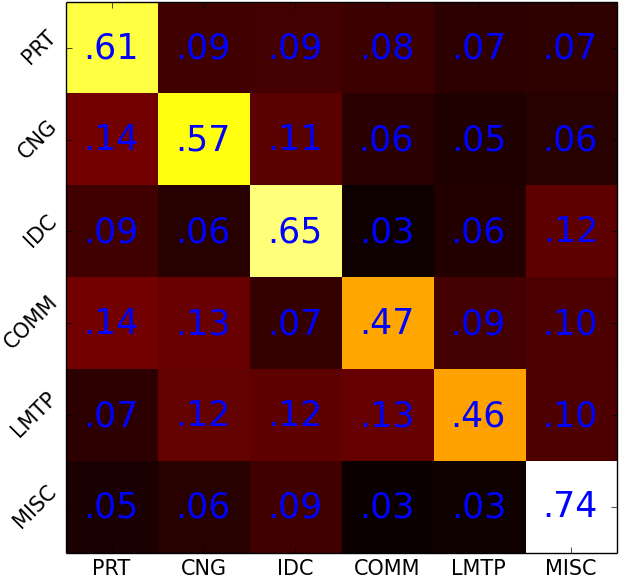}}\\%\vspace{-8pt}
  \subfigure[\scriptsize Hawkes process]{\label{fig:hawkes_confusion}
  \includegraphics[width=0.145\textwidth]{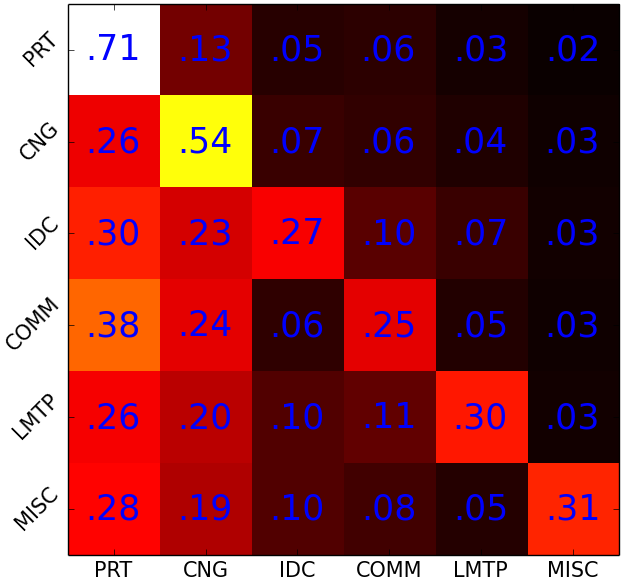}}%\hspace{-2pt}
  \subfigure[\scriptsize Logistic]{\label{fig:logit_confusion}
  \includegraphics[width=0.145\textwidth]{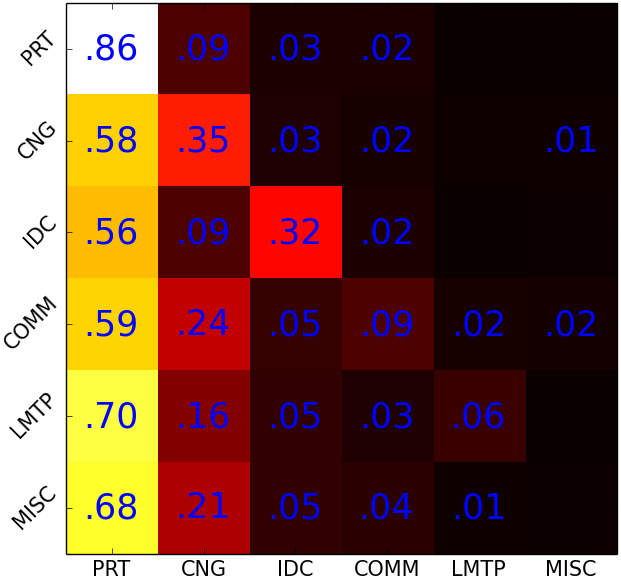}}%\hspace{-2pt}
  \subfigure[\scriptsize RMTPP]{\label{fig:rmtpp_confusion}
  \includegraphics[width=0.145\textwidth]{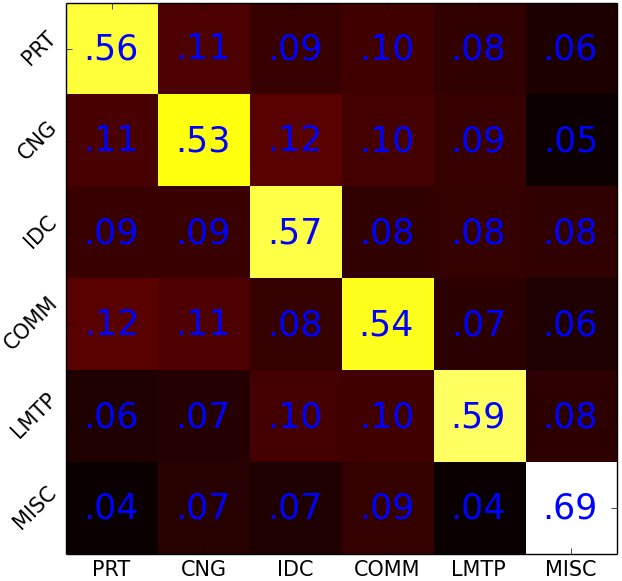}}\\%\hspace{-2pt}
  \subfigure[\scriptsize i-hRNN]{\label{fig:hiRNN_2confusion}
  \includegraphics[width=0.085\textwidth]{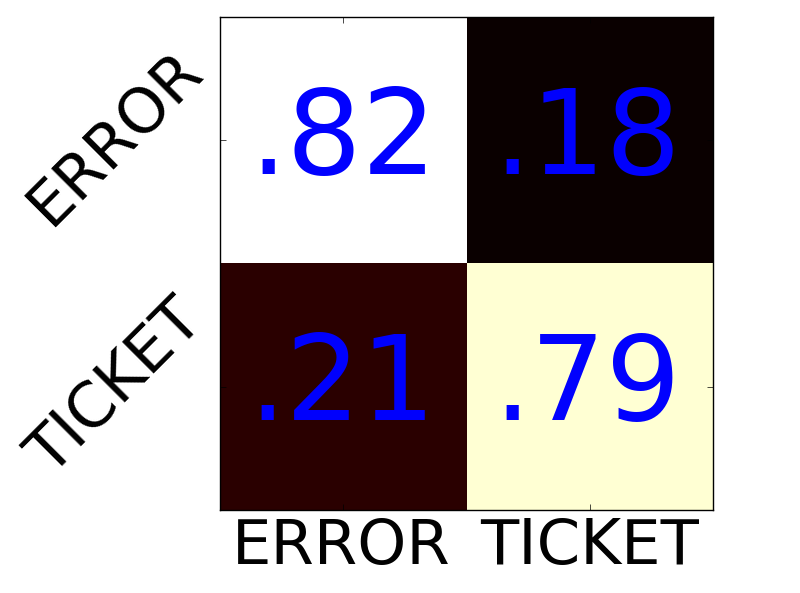}}%\hspace{-2pt}
  \subfigure[\scriptsize t-hRNN]{\label{fig:htRNN_2confusion}
  \includegraphics[width=0.085\textwidth]{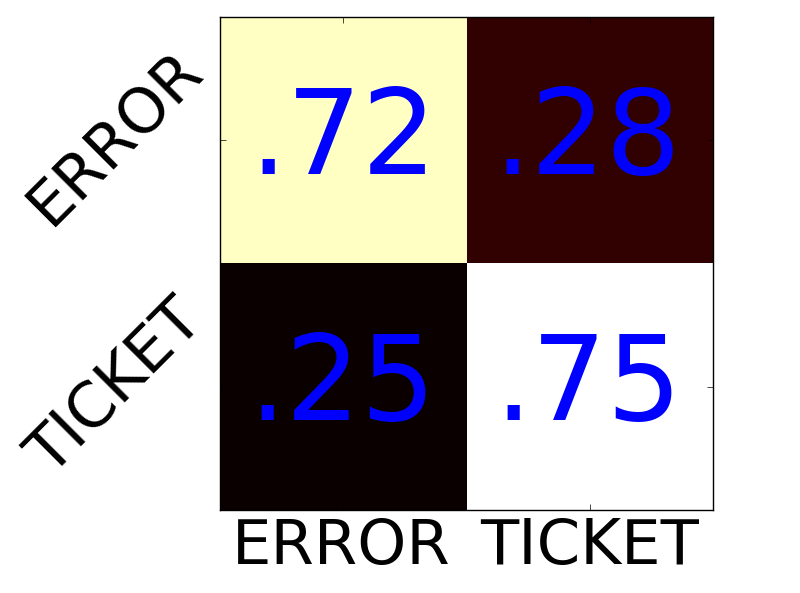}}%\vspace{-3pt}
  \subfigure[\scriptsize e-hRNN]{\label{fig:heRNN_2confusion}
  \includegraphics[width=0.085\textwidth]{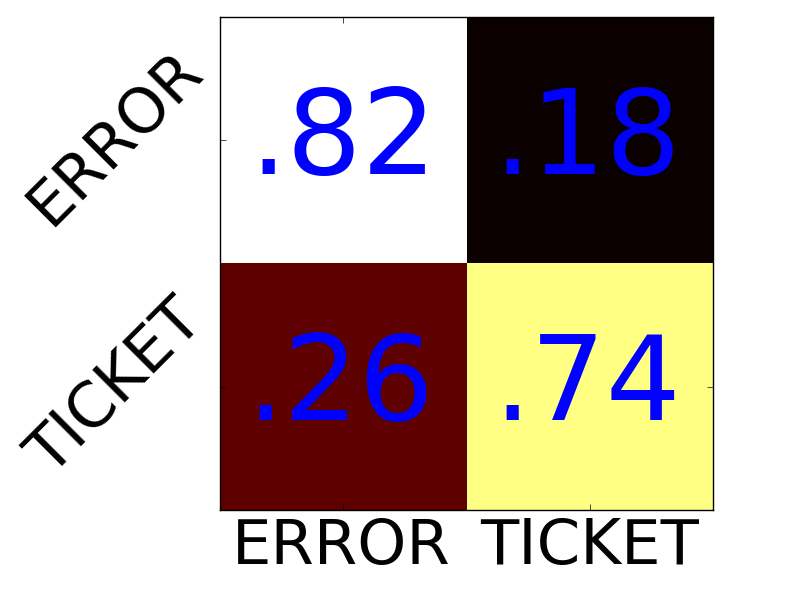}}%\hspace{-2pt}
  \subfigure[\scriptsize i-RNN]{\label{fig:iRNN_2confusion}
  \includegraphics[width=0.085\textwidth]{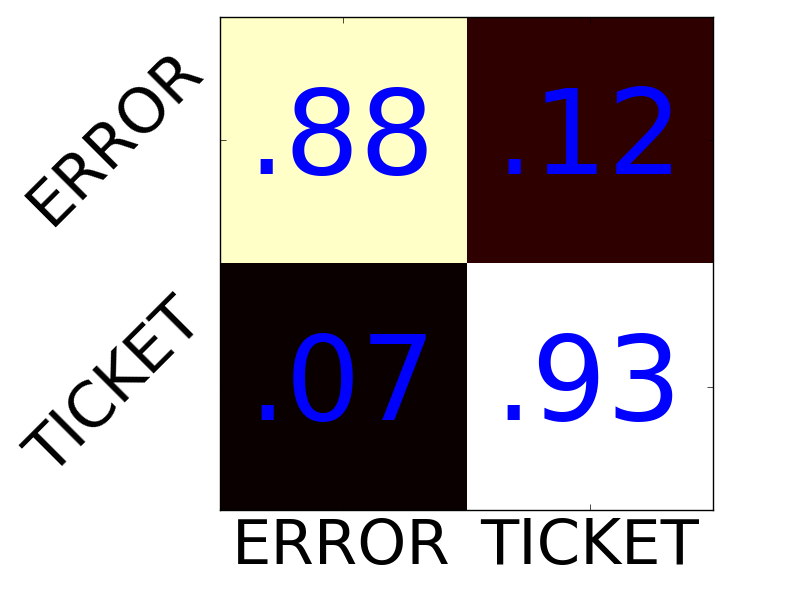}}%\hspace{-2pt}
  \subfigure[\scriptsize t-RNN]{\label{fig:tRNN_2confusion}
  \includegraphics[width=0.085\textwidth]{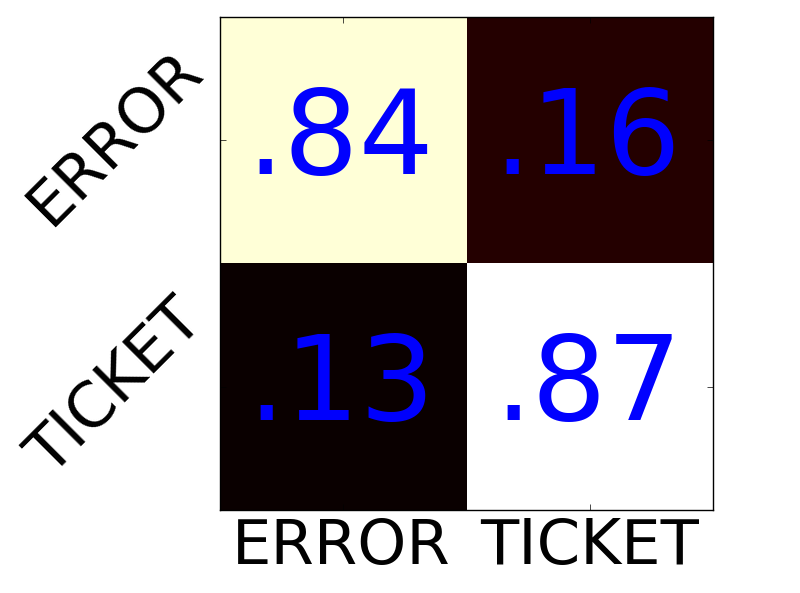}}\\%\vspace{-3pt}
  \subfigure[\scriptsize e-RNN]{\label{fig:eRNN_2confusion}
  \includegraphics[width=0.085\textwidth]{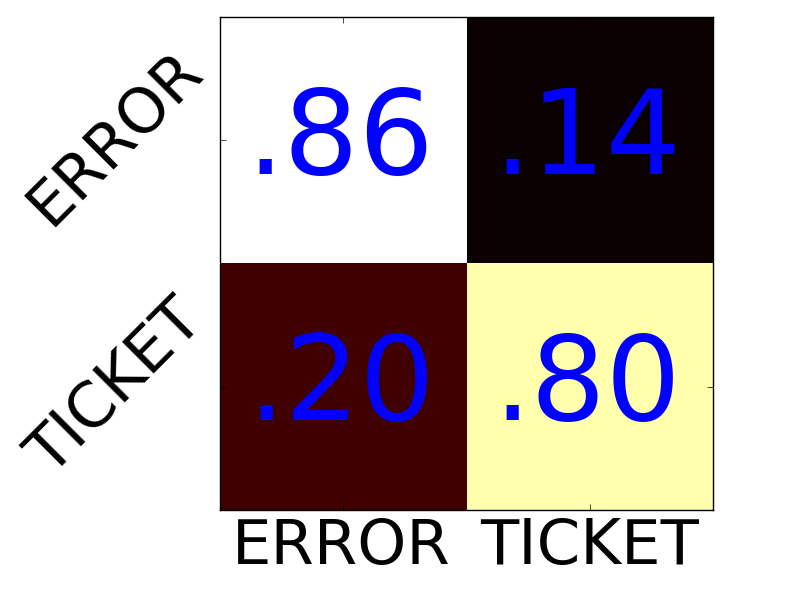}}
  \subfigure[\scriptsize Hawkes]{\label{fig:hawkes_2confusion}
  \includegraphics[width=0.085\textwidth]{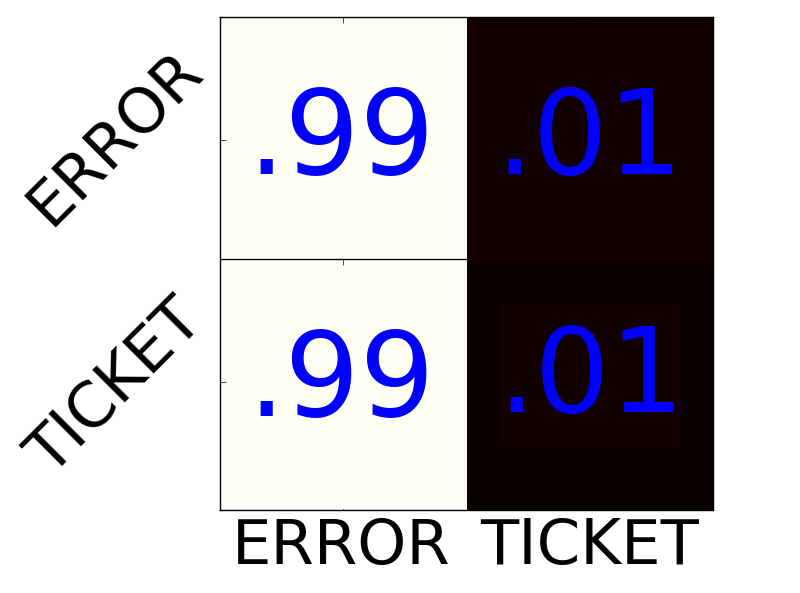}}%\hspace{-2pt}
  \subfigure[\scriptsize Logistic]{\label{fig:logit_2confusion}
  \includegraphics[width=0.085\textwidth]{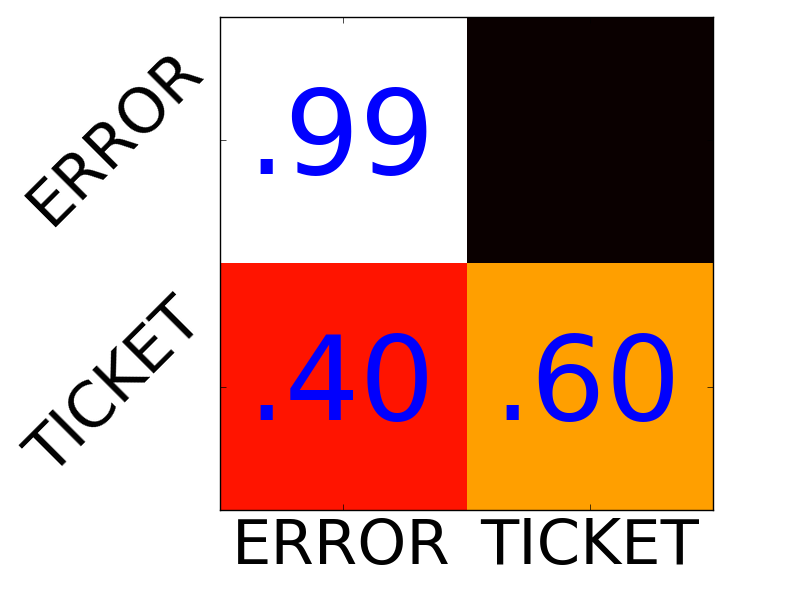}}%\hspace{-2pt}
  \subfigure[\scriptsize RMTPP]{\label{fig:rmtpp_2confusion}
  \includegraphics[width=0.085\textwidth]{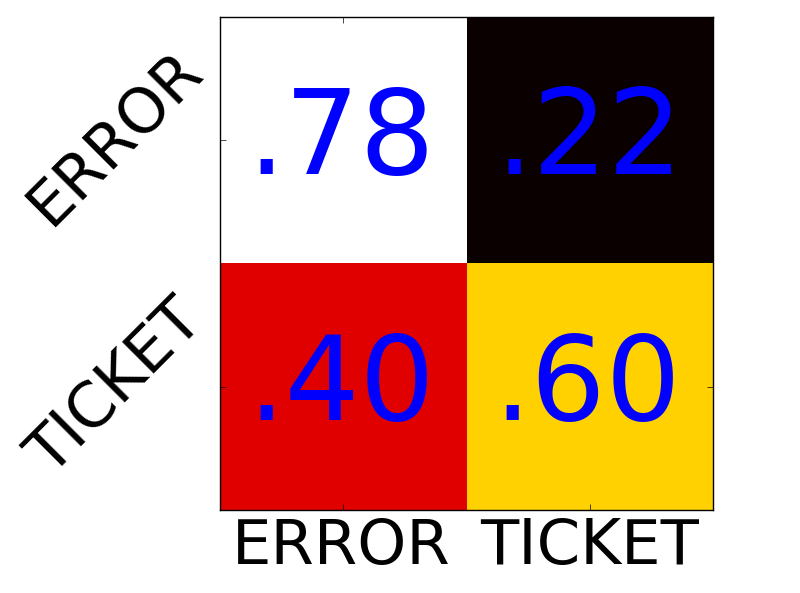}}%\vspace{-10pt}
	\caption{Confusion matrixes for sub/main type. Three methods in top and middle row use the hierarchical and flat structure as in Fig.\ref{fig:hierarchical} respectively. Zoom in for better view.}
	\label{fig:confusion}
\end{figure}

\subsection{Results and discussion}
All evaluations are performed on the testing dataset distinctive to training set whose statistics are shown in Table \ref{tab:type_statistics}.

\textbf{Averaged performance} Table \ref{tab:performance} shows the averaged performance among various types of events. As shown in Fig.\ref{fig:hierarchical}, we test two architectures of the event type prediction layer, i.e. hierarchical predictor (Fig.\ref{fig:hier_pred}) and flat independent predictors (Fig.\ref{fig:flat_pred}). The main type includes `ticket' and `error' and the subtype include `ticket' and the other six subtypes under `error' as we describe earlier in the paper.

\textbf{Confusion matrix} The confusion matrix for the six subtypes under `error' event, as well as for the two main types `ticket' and `error' are shown in Fig.\ref{fig:confusion} by various methods.

We make observations and analysis based on the results:

1) As shown in \ref{tab:performance}, for main-type, the flat architecture that directly predicts the main types outperforms the hierarchical one in different settings of the input RNN as well as varying evaluation metrics. This can be explained that the loss function focuses on the main-type misclassification only. While for the subtype prediction, the hierarchical layer performs better since it fuses the output from the main-type prediction layer and the embedding layer as shown in Fig.\ref{fig:hier_pred}.

2) No surprisingly, for both event type and timestamp prediction, our main approach, i.e. intensity RNN that fuses two RNNs outperforms its counterparts time series RNN and event sequence RNN by a notable margin. While the event RNN also often performs better than the time series counterpart. This suggests at least in the studied dataset, history event effects are important for the future event occurrence.

3) Our main method intensity RNN is almost always superior against other methods except for the main-type prediction task, whereby the Logistic classification model performs better. However for more challenging tasks i.e. subtype prediction and event timestamp prediction, our method significantly outperforms especially for subtype prediction task. Interestingly, all point process based models obtain better results on this task which suggests the point process models are more promising compared with classical classification models. Indeed, our methodology provides an end-to-end learning mechanism without any pre-assumption in modeling point process. All these empirical results on real-world tasks suggest the efficacy of our approach.

\begin{figure}[tb!]
\centering
  \includegraphics[width=0.48\textwidth]{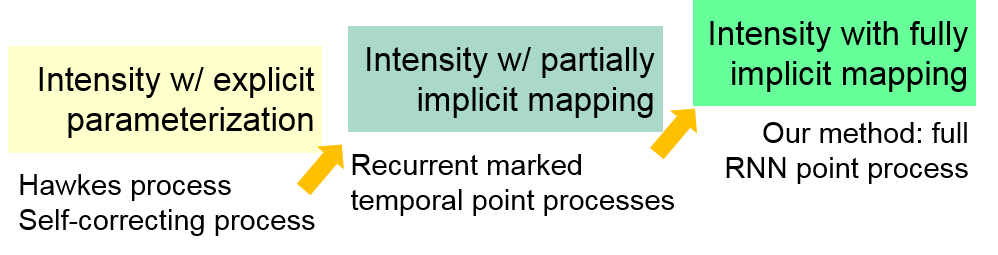}%\vspace{-10pt}
	\caption{The evolving of point process modeling.}
	\label{fig:evolve}
\end{figure}
\section{Conclusion}
We use Fig.\ref{fig:evolve} to conclude and further position our model in the development of (implicit and explicite) modeling the intensity function of point process. In fact, Hawkes process uses a full explicit parametric model and RMTPP misses the dense time series features to model time-varying base intensity and assumes a partially parametric form for it. We make a further step by a full implicit mapping model. Our model (see Fig.\ref{fig:overview}) is simple, general and can be learned end-to-end with standard backpropagation and opens up new possibilities for borrowing the advances in neural network learning to the area of point process modeling and applications. The representative study in this paper has clearly suggests its high potential to real-world problems, even we have no domain knowledge on the problem at hand. This is in contrast to existing point process models where an assumption about the dynamics is often need to be specified beforehand.
\small
\bibliographystyle{aaai}
\bibliography{xiao-yan}
\end{document}